\documentclass[journal,twoside,web]{ieeecolor}
\usepackage{generic}
\usepackage{cite}
\usepackage{amsmath,amssymb,amsfonts}
\usepackage{algorithmic}
\usepackage{graphicx}
\usepackage{textcomp}
\def\BibTeX{{\rm B\kern-.05em{\sc i\kern-.025em b}\kern-.08em
    T\kern-.1667em\lower.7ex\hbox{E}\kern-.125emX}}

\usepackage{amsmath,amsfonts,amssymb}
\usepackage{pdfpages}
\usepackage[bookmarks=false,colorlinks=true,citecolor=blue,linkcolor=blue]{hyperref}
\usepackage{mathtools}
\usepackage{bbm}
\usepackage{epsfig} %
\usepackage{times} %
\usepackage{xcolor}
\usepackage[english]{babel}
\usepackage{graphicx}
\usepackage{subfigure}
\usepackage{nicefrac}

\def\argmin{\operatornamewithlimits{arg\,min}}

\newcommand{\defeq}{\doteq}
\newcommand{\gramdplus}{\widetilde{K}_{0}}
\newcommand{\gramnplus}{K_{0}}
\newcommand{\Ker}{K}
\newcommand{\newpart}{}

\newcommand{\BE}{\mathbb{E}}

\newcommand{\BX}{\mathbb{X}}

\newcommand{\BI}{\mathbb{I}}

\newcommand{\CL}{\mathcal{L}}

\newcommand{\CH}{\mathcal{H}}

\makeatletter
\newcommand*{\rom}[1]{\expandafter\@slowromancap\romannumeral #1@}
\makeatother

\def \CD{\mathcal{D}}

\def \RKHS{\mathcal{H}}
\def \Ker{K_x}

\def \Ker{K}

\DeclarePairedDelimiter\ceil{\lceil}{\rceil}

\newcommand{\tr}{^\mathrm{T}}  %
\newcommand{\RR}{\mathbb{R}}

\def\argmin{\operatornamewithlimits{arg\,min}}        %

\newtheorem{assumption}{\bf A\!\!}
\newtheorem{theorem}{Theorem}
\newtheorem{lemma}{Lemma}

\newtheorem{remark}{Remark}

\newcommand{\norm}[1]{\left\lVert#1\right\rVert}    
    
\newlength{\dhatheight}

\begin{document}

\title{\LARGE \bf
Nonparametric, Nonasymptotic {\newpart Confidence Bands} with\\
Paley-Wiener Kernels for Band-Limited Functions}
\author{Bal{\'a}zs Csan{\'a}d Cs{\'a}ji, \IEEEmembership{Member, IEEE}, \and B\'alint Horv\'ath
\thanks{\newpart This research was supported by the National Research, Development and Innovation Office (NRDIO) of Hungary, 
within the framework of the Artificial Intelligence National Laboratory Program; and by 
the Thematic Excellence Programme (TKP) of NRDIO, pr.~no.~TKP2021-NKTA-01.}%
\thanks{\newpart B.~Cs.~Cs\'aji is with SZTAKI: Institute for Computer Science and Control, Budapest, Hungary; and also with Institute of Mathematics, E\"otv\"os Lor\'and University (ELTE), Budapest, Hungary, {\tt\small csaji@sztaki.hu}}%
\thanks{\newpart B.~Horv\'ath is with SZTAKI: Institute for Computer Science and Control, E\"otv\"os Lor\'and Research Network (ELKH), Budapest, Hungary; and also with  Institute of Mathematics, Budapest University of Technology and Economics (BME), Hungary, {\tt\small balint.horvath@sztaki.hu}}%
}

\hyphenation{pa-ram-et-ri-za-ti-on}

\maketitle
\thispagestyle{empty}
\pagestyle{empty}

\begin{abstract}
The paper introduces a {\newpart method} to construct {\newpart confidence bands} for bounded, band-limited functions based on a finite sample of input-output pairs. The {\newpart approach} is distribution-free w.r.t.\ the observation noises {\newpart and only the knowledge of the input distribution is assumed.} It is nonparametric, that is, it does not require a parametric model of the regression function {\newpart and} the regions have non-asymptotic guarantees. The {\newpart algorithm} is based on the theory of Paley-Wiener reproducing kernel Hilbert spaces. The paper first studies the fully observable variant, when there are no noises on the observations {\newpart and} only the inputs are random{\newpart ;} then {\newpart it} generalizes the ideas to the noisy case using {\newpart gradient-perturbation} methods. Finally, numerical experiments demonstrating both cases are presented.
\end{abstract}


\section{Introduction} 
\IEEEPARstart{R}{\lowercase{egression}} is one of the fundamental problems of statistics, system identification, signal processing and machine learning {\newpart  \cite{cucker2007learning}}. Given a finite sample of input-output pairs, the typical aim is to estimate 
the so-called {\em regression function}{\newpart , which, given an input, encodes the conditional expectation of the corresponding output} \cite{ljung2010perspectives}. There are several well-known (parametric and nonparametric) approaches for regression, from linear regression to neural networks and kernel methods, which provide {\em point-estimates} from a {\newpart given} model class
\cite{gyorfi2002distribution}.

However, sole point-estimates are often not sufficient and {\em region-estimates} are also needed, for example,
to support {\em robust} approaches. These region-estimates have several variants, such as {\em confidence regions} for the ``true'' function generating the observations \cite{Algo2018}; for the {\em expected} 
output at a given input \cite{quinonero2005unifying}; and {\em prediction regions} for the next (noisy) observation \cite{vovk2005algorithmic}.

In this paper, we focus on building {\em confidence bands} for the regression function. These bands have natural connections to filtering and smoothing methods.
While in a {\em parametric} setting such region-estimates are typically induced by confidence sets
in the parameter space, in a {\em nonparametric} setting this indirect approach is not feasible. 
Therefore, nonparametic confidence bands for the expected outputs should be constructed directly.

Regarding prediction intervals for the {\em next observation}, promising distribution-free approaches are {\em interval predictor models} (IPMs) based on the scenario approach \cite{campi2009interval, garatti2019class}, and the {\em conformal prediction} framework also offers {\newpart  several nonparametric methods for regression and classification} \cite{vovk2005algorithmic}. 

{\newpart If} the data is jointly Gaussian, a powerful methodology is offered by {\em Gaussian process regression} \cite{quinonero2005unifying} that can provide 
prediction regions for the outputs, and {\newpart  credible regions for the expected outputs}. However, the Gaussianity assumption is sometimes unrealistic that calls for alternative
approaches.

In this paper, we suggest a {\em nonparametric} approach using Paley-Wiener kernels,
to build data-driven {\em simultaneous} confidence bands for an unknown bounded, {\em band-limited} function, based on an independent and identically distributed (i.i.d.) sample of input-output pairs. The method is {\em distribution-free} in the sense that only very mild assumptions are needed about the observation noises, such as they are distributed {\em symmetrically} about zero. On the other hand, we assume that the {\em distribution of the inputs} is known, particularly, we 
assume uniformly distributed inputs, as more general cases can often be traced back to this assumption. First,
the case without observation noises is studied,
then the ideas are extended to the general, noisy case. The results are supported by both {\em non-asymptotic} theoretical guarantees and numerical experiments.

\section{Kernels and Band-Limited Functions}
{\newpart Kernel methods have an immerse range of applications in machine learning and related fields \cite{pillonetto2014kernel}. 
In this section, we review some of their fundamental theoretical concepts.}

\subsection{Reproducing Kernel Hilbert Spaces}
A Hilbert space $\CH$ of $f: \BX \to \mathbb{R}$ functions with an inner product $\langle\cdot,\cdot\rangle_{\CH}$ is called a {\em Reproducing Kernel Hilbert Space} (RKHS), if each Dirac functional, which evaluates functions at a point,
$\delta_z: f \to f(z)$, is 
bounded for all $z \in \BX$, that is $\forall z \in \BX: \exists \, \kappa_z > 0$ with $|\hspace{0.3mm}\delta_z(f)\hspace{0.3mm}| \leq \kappa_z\, \| f \|_{\CH}$ for all $f \in \CH$.

Then, by building on the Riesz representation theorem, a unique {\em kernel}, $k: \BX \times \BX \to \mathbb{R}$,  can be constructed %
encoding the Dirac functionals satisfying $\langle k(\cdot,z),f \rangle_{\CH} = f(z),$
for all 
$z \in \BX$ and $f \in \CH$, which formula is called the {\em reproducing property}.
As a special case of this property, we also have for 
all $z,s \in \BX$ {\newpart that} $k(z,s)=\langle k(\cdot,z),k(\cdot,s) \rangle_{\CH}.$ 
Therefore, 
the kernel of an RKHS is a symmetric and positive-definite function.

Furthermore, the Moore-Aronszajn theorem asserts that the converse statement
holds true, as well: 
for every symmetric and positive-definite function $k: \BX \times \BX \to \mathbb{R}$, there exists a unique RKHS for which $k$ is its reproducing kernel \cite{berlinet2004reproducing}.

The {\em Gram} or kernel matrix of a given kernel $k$ w.r.t.\ (input) points $x_1, \dots, x_n$ is 
$K_{i,j} \defeq k(x_i,x_j)$, for all {\newpart  $i, j \in [n] \doteq \{1,\dots, n\}$}. Observe that $K \in \mathbb{R}^{n \times n}$ is always positive semi-definite. A kernel is called {\em strictly} positive-definite, if its Gram matrix is 
positive-definite for all {\em distinct} inputs $\{x_i\}$.

{\newpart  Archetypal} kernels include the Gaussian kernel $k(z,s)=\exp (-||z-s||^2 / (2 \sigma^2)),$ where $\sigma >0$; the polynomial kernel $k(z,s)=(\langle z,s \rangle +c)^p,$ where 
$c \geq 0$, $p \in \mathbb{N}$; 
and the sigmoidal kernel  $k(z,s)=\tanh (a \langle z,s \rangle +b),$ 
for some $a,b \geq 0$.

\subsection{Paley-Wiener Spaces}
Let  $\CH$ be the space of  $f \in \CL^2 (\mathbb{R}, \lambda)$ functions, where $\lambda$ is the Lebesgue measure, such that the support of the {\em Fourier transform} of $f$ is included in $[\hspace{0.3mm}-\eta,\, \eta\hspace{0.5mm}]$, where $\eta > 0$. It is a subspace of $\CL^2$ and thus we use the $\CL^2$ inner product:\vspace{-0.5mm}
$$\langle f, g \rangle_\CH \, \defeq \int_{\mathbb{R}} f(x)\,g(x) \: \mathrm{d} \lambda(x).$$

This space of {\em band-limited} functions, called the {\em Paley-Wiener space} \cite{berlinet2004reproducing}, is an RKHS.
Its reproducing kernel is
$$k(z,s) \, \defeq \, \frac{\sin (\eta(z-s))}{\pi(z-s)},$$
for $z \neq s$, where $z, s \in \mathbb{R}$; and $k(z, z) \defeq \eta/\pi$.
Henceforth, we will work with the above defined {\em Paley-Wiener kernel}.

\begin{remark}
Paley-Wiener spaces can also be defined on $\mathbb{R}^d$ \cite{iosevich2015exponential}, but 
for simplicity we focus on the scalar input case.
\end{remark}

\section{Nonparametric {\newpart Confidence Bands}}

Let $(x_1, y_1), \dots, (x_n, y_n)$ be a finite sample of i.i.d. pairs of 
random variables with unknown joint distribution $\mathbb{P}_{\! \scriptscriptstyle X,Y}$, where 
$x_k$ and $y_k$ are $\RR$-valued, and 
$\mathbb{E}[\hspace{0.3mm}y^2_k\hspace{0.3mm}] < \infty$. We assume that\vspace{-0.5mm}
$$
y_k \, = \, f_*(x_k) + \varepsilon_k,
$$
for $k \in [n]$, where $\mathbb{E}[\hspace{0.3mm}\varepsilon_k\hspace{0.3mm}] = 0$.
Variables $\{\varepsilon_k\}$
represent 
the measurement or observation {\em noises} {\newpart  on the ``true'' $f_*$.}

We call $f_*$ the {\em regression function}\hspace*{-0.5mm} {\newpart \cite{cucker2007learning}}, as on the support of $\{x_k\}$ it can also be written as
$
f_*(x) \,= \, \mathbb{E} \left[\hspace{0.5mm} Y\hspace{0.5mm} |\hspace{0.5mm} X = x \hspace{0.5mm}\right]
$,
where $(X,Y)$ is a
random vector with distribution $\mathbb{P}_{\! \scriptscriptstyle X,Y}$.
\subsection{Objectives and Reliability}
\label{sec:objectives}
Our aim 
is to {\newpart build a (simultaneous) {\em confidence band}} for $f_*$, i.e., a function $I:\mathcal{D} \to {\newpart \mathbb{R} \times \mathbb{R}}$, where $\mathcal{D}$ is the {\em support} of the input distribution, 
such that {\newpart $I(x) = (\hspace{0.3mm}I_1(x), I_2(x)\hspace{0.3mm})$ specifies the {\em endpoints} of an interval estimate for $f_*(x)$, for all $x \in \mathcal{D}$}.
More precisely, 
we would like to construct $I$ with\vspace{-0.5mm}
$$
\nu(I)\,\defeq \, \mathbb{P} \big(\, \forall x \in \mathcal{D}: {\newpart I_1(x) \leq f_*(x) \leq I_2(x)} \,\big) \, \geq \, 1- \alpha,
$$
where $\alpha \in (0,1)$ is a user-chosen {\em risk} probability, and $\nu(I)$ is 
{\newpart the {\em reliability} of the confidence band.
Let us introduce}
\vspace{-0.2mm}
$$
\mathcal{I} \, \defeq \, \big\{\hspace{0.5mm}   (x,y) \in \CD \times \mathbb{R} : y \in [ \hspace{0.3mm} I_1(x), I_2(x) \hspace{0.3mm} ] \hspace{0.5mm}   \big\}.
$$
{\newpart Based} on this, the reliability is $\nu(I) = \mathbb{P}(\, \mathrm{graph}_{\CD}(f_*) \subseteq \mathcal{I}\,)$, where we define $\mathrm{graph}_{\CD}(f_*) \defeq \{\, (x, f_*(x)) : x\in \CD \,\}$.

{\newpart For notational simplicity, we will use $I(x) = \emptyset$ to denote $I(x) = (\hspace{0.3mm}1,-1\hspace{0.3mm})$, i.e., the endpoints of an empty interval.}

Hence, we aim at building a {confidence band} that contains the graph (w.r.t.\ domain $\CD$) of the ``true'' $f_*$ 
with a {\em user-chosen} probability {\newpart  level}. Moreover, we would like to have a {\em distribution-free} method (w.r.t.\ the noises) and the region should have {\em finite-sample} guarantees without a parametric model of $f_*$, namely, we take a {\em nonparametric} approach.

\begin{remark}
We 
note here, as well, that in the 
IPMs \cite{campi2009interval}\cite{garatti2019class} and in the conformal prediction framework \cite{vovk2005algorithmic}, the aim is to build a guaranteed prediction region for the {\em next observation}, while here we aim at predicting the value of the {\em regression function} instead. In this sense, {\newpart our objective is similar to that of the region estimates of Gaussian process regression} \cite{quinonero2005unifying}, however, without the assumption {\newpart of joint Gaussianity}.
\end{remark}

\subsection{Main Assumptions}

Our core  assumptions can be summarized as follows:
\smallskip

 \setcounter{assumption}{-1}

\begin{assumption}
\label{A0} %
{\em The dataset, $(x_1, y_1), \dots, (x_n, y_n) \in \RR \times \RR$, is an i.i.d.\ sample of input-output pairs; and $\mathbb{E}[\hspace{0.3mm}y^2_k\hspace{0.3mm}] < \infty$, for $k \in [n]$}.
\end{assumption}
\smallskip

\begin{assumption}
\label{A1} {\em Each (measurement) noise, $\varepsilon_k \doteq y_k - f_*(x_k)$,  for $k \in [n]$, has a {symmetric} probability distribution about zero.}
\end{assumption}
\smallskip

\begin{assumption}
\label{A2} {\em The inputs, $\{x_k\}$, are distributed uniformly on $[\hspace{0.4mm}0, 1\hspace{0.2mm}]$.}
\end{assumption}
\smallskip

\begin{assumption}
\label{A3} {\em Function 
$f_*$ is from a Paley-Wiener space $\CH$;
$\forall\,  x\in[\hspace{0.4mm}0, 1\hspace{0.2mm}]: {\newpart |f_*(x)|} \leq 1$; and
$f_*$ is almost time-limited to $[\hspace{0.4mm}0, 1\hspace{0.3mm}]:$
$$ 
 \int_{\mathbb{R}} f^2_*(x)\,\BI(x \notin  [\hspace{0.4mm}0, 1\hspace{0.2mm}]) \: \mathrm{d}\lambda(x) \, \leq \, \delta_0,
$$ 
where $\BI(\cdot)$ is an indicator and $\delta_0 > 0$ is a universal constant.}
\end{assumption}
\smallskip

Now, let us briefly discuss these assumptions. The i.i.d.\ requirement of A\ref{A0} is standard in mathematical statistics and supervised learning \cite{Vapnik1998}.  
The square-integrability of the outputs is needed to estimate the $\CL^2$ norm of $f_*$ based on the sample and to have a well-defined regression function. 
The assumption on the 
noises, A\ref{A1}, is very mild, as most standard distributions (e.g., Gauss, Laplace and uniform) satisfy this.

Our strongest assumption is certainly A\ref{A2}, 
which basically {\newpart amounts} to the assumption that {\em we know the distribution of the inputs} and it is absolutely continuous. The more general case when the inputs, $\{x'_k\}$, have a {\em known}, strictly monotone {\newpart increasing} and continuous cumulative distribution function $F$, could be traced back to assumption A\ref{A2}, {\newpart since} it is well-known that $x_k \defeq F(x'_k)$ is distributed uniformly on $[\hspace{0.4mm}0, 1\hspace{0.2mm}]$. 

Assumption A\ref{A3}, especially limiting the frequency domain of $f_*$,
is needed to restrict the model class and to ensure that we can effectively generalize to unknown data points. We allow 
the ``true'' function to be defined outside the support of the inputs, cf.\ the Fourier uncertainty principle{\newpart \cite{pinsky2008introduction}}, but the part of $f_*$ outside of $\CD = [\hspace{0.4mm}0, 1\hspace{0.2mm}]$ should be ``negligible'', i.e., its norm cannot exceed a {\newpart (known)} small constant, $\delta_0$. 

{\newpart A crucial property of Paley-Wiener spaces is that their norms coincide with the standard $\mathcal{L}^2$ norm, which will allow us to efficiently upper bound $\|f_*\|_{\CH}^2$ based on the sample.}

\section{{\newpart Confidence Bands}: Noise-Free Case}
In order to motivate our solution, we start with a simplified problem, in which we observe the regression function perfectly at 
random inputs. In this noise-free case, we can recall the celebrated Nyquist–Shannon sampling theorem, which states that a band-limited function can be fully reconstructed from the samples, assuming the sampling rate exceeds twice the maximum frequency. On the other hand, if we only have a small number of observations, we cannot apply this result. Nevertheless, we still would like to have at least a region estimate. In this section we provide such an algorithm.
Recall that for a dataset $\{(x_k, y_k)\}$, where inputs $\{x_k\}$ are {\em distinct} (which has probability one under A\ref{A2}), the element from $\CH$ that has the {\em minimum norm} and {\em interpolates} each output $y_k$ at the corresponding input $x_k$, that is\vspace{-0.5mm}
$$
\bar{f} \, \defeq \, \argmin \big\{\,\|\hspace{0.3mm}f\hspace{0.4mm}\|_{\CH} : f \in \CH\hspace{1.5mm} \&\hspace{1.5mm} \forall\hspace{0.3mm} k \in [n]: f(x_k) =\, y_k   \,  \big\},\vspace{-0.5mm}
$$
takes the following form \cite{berlinet2004reproducing} for all input $x \in \BX:$\vspace{-0.5mm}
$$\bar{f}(x)\,=\, \sum_{k=1}^n \bar{\alpha}_k k(x, x_k),\vspace{-0.5mm}$$
where the weights are $\bar{\alpha} = K^{-1} y$ with $y\defeq (y_1, \dots, y_n)\tr$ and $\bar{\alpha} \defeq (\bar{\alpha}_1, \dots, \bar{\alpha}_n)\tr$; we also used that the Paley-Wiener kernel is strictly positive-definite, thus 
matrix $K$ is invertible.

We will exploit, as well, that the norm square of $\bar{f}$ is\vspace{-0.5mm}
$$\|\hspace{0.3mm}\bar{f}\hspace{0.4mm}\|_{\CH}^2 = \bar{\alpha}\tr \hspace{-0.3mm}K \bar{\alpha},\vspace{-0.5mm}$$
which is a direct consequence of the reproducing property.

Assuming we have a stochastic upper bound for the norm square of the regression function, denoted by $\kappa$, the idea of our construction is as follows. We include those $(x_0,y_0)$ pairs in the {\newpart confidence band}, for which the minimum norm interpolation of $\{(x_k, y_k)\} \,\cup\, \{(x_0,y_0)\}$, namely, which simultaneously interpolates the original dataset and $(x_0,y_0)$, has a norm square which is less than or equal to $\kappa$. In order to make this approach practical, we need (1) a guaranteed upper bound for the norm square of the {\newpart ``true''} data-generating function; and (2) an efficient method to decide the endpoints of the {\newpart confidence} interval for each potential input $x_0 \in \CD$.

%

%

%
%
\subsection{Bounding the Norm: Noise-Free Case}
It is easy to see that in the noise-free case, if $y_k = f_*(x_k)$, for $k \in [n]$, the norm square of $f_*$ can be estimated by \vspace{-0.5mm}
$$\frac{1}{n} \sum_{k=1}^n y_k^2 = \frac{1}{n} \sum_{k=1}^n f_*^2(x_k) \approx \mathbb{E}\big[ f^2_*(X)\big] \approx \|\hspace{0.3mm}f_*\hspace{0.4mm}\|_{2}^2 = \|\hspace{0.3mm}f_*\hspace{0.4mm}\|_{\CH}^2,$$
since in the Paley-Wiener space the norm is the $\CL^2$ norm, and we also used that $\{x_k\}$ are uniform on domain
$\CD = [\hspace{0.4mm}0, 1\hspace{0.2mm}]$.

As the next lemma demonstrates, we can construct such a guaranteed upper bound using the Hoeffding inequality:
\medskip
\begin{lemma}
\label{lemma:Hoeffding.noiseless}
{\em Assuming A\ref{A0}, A\ref{A2}, A\ref{A3} and that $y_k = f_*(x_k)$, for $k \in [n]$, 
{\newpart we have for any risk probability $\alpha\in (0,1)$,\vspace{-0.5mm}
$$
\mathbb{P}\big(\norm{f_*}_{\CH}^2 \leq \kappa \hspace{0.3mm}\big) \, \geq \, 1-\alpha,
$$
with the following choice of the upper bound $\kappa$:\vspace{-0.5mm}
$$
\kappa \, \defeq\, \frac{1}{n} \sum_{k=1}^n y_k^2 + \sqrt{\frac{\ln(\alpha)}{-2n}} + 
\delta_0.$$}
}	
\end{lemma}
\vspace{-1.5mm}
\hspace*{-8mm}
\begin{proof}
By using the notation ${\newpart R} \defeq \nicefrac{1}{n}\sum_{k=1}^n y_k^2$, we have 
$$\BE[\hspace{0.3mm} {\newpart R}\hspace{0.3mm} ]\, =\, \|\hspace{0.3mm} f_* \cdot \mathbb{I}_{\CD} \hspace{0.3mm} \|_2^2\, \geq\, \|\hspace{0.3mm}f_*\hspace{0.4mm}\|_{\CH}^2 - \delta_0,
$$
where $\mathbb{I}_{\CD}$ is the indicator function of $\CD = [\hspace{0.4mm}0, 1\hspace{0.2mm}]$. That is, ${\newpart R}$ is a Monte Carlo estimate of the integral of this $\CL^2$ norm.

Then, from the Hoeffding inequality, for all $t>0$:
$$\mathbb{P}({\newpart R} - \mathbb{E}[\hspace{0.3mm} {\newpart R}\hspace{0.3mm} ] \leq -t) \leq \mbox{exp} (-2n t^2).$$
According to the complement rule, we also have
$$\mathbb{P} ( \mathbb{E}[\hspace{0.3mm} {\newpart R}\hspace{0.3mm}]  < {\newpart R} + t) \geq 1-\mbox{exp}(-2nt^2).$$
We would like choose a threshold $t > 0$ such that
$$1-\alpha \, \leq\, \mathbb{P} ( \mathbb{E}[\hspace{0.3mm} {\newpart R}\hspace{0.3mm}] < {\newpart R}+t).$$
{\newpart This} inequality is satisfied if we choose a $t>0$ with
$$1-\alpha \leq 1-\mbox{exp}(-2nt^2)\; \Longrightarrow \;\mbox{exp}(-2nt^2) \leq \alpha.$$
After taking the natural logarithm, we get 
$-2nt^2 \leq \ln(\alpha)$, 
hence, the choice of 
$t^* = \sqrt{\ln(\alpha)/(-2n)}$
guarantees
$$\mathbb{P}\big( \hspace{0.3mm} \|\hspace{0.3mm}f_*\hspace{0.4mm}\|_{\CH}^2 \geq {\newpart R} +t^*+\delta_0  \hspace{0.3mm} \big) \leq \alpha,$$
which completes the proof of the lemma.
\end{proof}

\smallskip
\subsection{Interval Endpoints: Noise-Free Case}
Now, we construct a {\newpart confidence} interval for a given input {\em query point} $x_0 \in \CD$, for which $x_0 \neq x_k$, for $k \in [n]$. That is, we build an {\newpart interval $[I_1(x_0),I_2(x_0)]$} that contains $f_*(x_0)$ with probability at least $1-\alpha$, where $\alpha \in (0,1)$ is given.

First, we extend the Gram matrix with query point $x_0$,
$$
\gramnplus({i+1},{j+1})\, \defeq \, k(x_i,x_j),
$$
for $i, j = 0,1, \dots ,n$. As $\{x_k\}_{k=0}^n$ are distinct (a.s.), this Gramian can
be inverted. Hence, for any $y_0$, the minimum norm interpolation of $(x_0, y_0), (x_1, y_1), \dots, (x_n, y_n)$ is \vspace{-0.5mm}
$$\tilde{f}(x)\,=\, \sum_{k=0}^n \tilde{\alpha}_k k(x, x_k),$$
where the weights are $\tilde{\alpha} = \gramnplus^{-1} \tilde{y}$ with $\tilde{y}\defeq (y_0, y_1, \dots, y_n)\tr$ and $\tilde{\alpha} \defeq (\tilde{\alpha}_0, \dots, \tilde{\alpha}_n)\tr.$ 
The norm square of $\tilde{f}$ is
$$
\|\hspace{0.3mm}\tilde{f}\hspace{0.4mm}\|_{\CH}^2 \,=\, \tilde{\alpha}\tr\hspace{-0.3mm} \gramnplus \tilde{\alpha}\, =\, \tilde{y}\tr\hspace{-0.3mm} \gramnplus^{-1} \gramnplus \gramnplus^{-1} \tilde{y}\,=\, \tilde{y}\tr\hspace{-0.3mm} \gramnplus^{-1} \tilde{y}.
$$
Since the output query point $y_0$ in $\tilde{y} = (y_0, y\tr)\tr$ is arbitrary, we can compute the minimum norm needed to interpolate the original dataset extended by $(x_0, y_0)$ for any 
candidate $y_0$. 

Therefore, having a bound $\kappa$ on the norm square (which is guaranteed with probability $\geq 1-\alpha$), we can compute the highest and the lowest $y_0$ values which can be interpolated with a function from $\CH$ having at most norm square $\kappa$. 

This leads to the following {\em two} optimization problems:
\begin{equation}
\label{noiseless-opt-min-max}
\begin{split}
\mbox{min\,/\,max} &\quad y_{0} \\[0.5mm]
\mbox{subject to} &\quad (y_0, y\tr)  \gramnplus^{-1} (y_0, y\tr)\tr \leq\, \kappa\\[1mm]
\end{split}
\end{equation}
where ``min\,/\,max'' means that we have to solve the problem as a minimization and also as a maximization (separately). 

The optimal values of these problems, denoted by $y_{\mathrm{min}}$ and $y_{\mathrm{max}}$, respectively, determine the {\em endpoints} of the {\newpart confidence} interval for $f_*(x_0)$, that is 
$I_1(x_0) \defeq y_{\mathrm{min}}$ and $I_2(x_0) \defeq y_{\mathrm{max}}$.

Problems  \eqref{noiseless-opt-min-max} are convex, moreover, as we will show, 
their optimal vales 
can be calculated {\em analytically}. First, note that the only decision variable of these problems is $y_0$, everything else is constant (including the input 
$x_0$, which is also given).

Let us partition the inverse Gramian, $\gramnplus^{-1}$, as\vspace{-0.2mm}
$$
\begin{bmatrix}
\; c & b\tr\\
\; b & A
\,\end{bmatrix} \defeq\, \gramnplus^{-1}\!\!,
$$
where $c \in \RR$, $b\in \RR^n$ and $A \in \RR^{n\times n}$; after which
$$
\quad (y_0, y\tr)  \gramnplus^{-1} (y_0, y\tr)\tr  =\, c\, y_0^2 + 2\, b\tr y\, y_0 + y\tr\hspace{-0.3mm} A y.
$$
Then, introducing $a_0 \defeq c$, $b_0 \defeq 2b\tr y$ and $c_0 = y\tr\hspace{-0.3mm} A y - \kappa$, the two optimization problems \eqref{noiseless-opt-min-max} can be written as 
\begin{equation}
\label{noiseless-opt-proof}
\begin{split}
\mbox{min\,/\,max} &\quad y_{0} \\[0.5mm]
\mbox{subject to} &\quad a_0 y_0^2 + b_0 y_0 + c_0 \, \leq \, 0
\end{split}
\end{equation}
in which $a_0$, $b_0$ and $c_0$ are constants (w.r.t.\ the optimization).

Since these are (convex) quadratic programming problems (with linear objectives), their optimal solutions must be on the boundary of the constraint. This can be easily verified directly, for example, by the technique of Lagrange multipliers.

There are at most two solutions of the quadratic equation  $a_0 y_0^2 + b_0 y_0 + c_0 = 0.$ 
The smaller one will be denoted by $y_{\mathrm{min}}$ and the larger one by $y_{\mathrm{max}}$ (they are allowed to be the same, if there is only one solution). 
Then, we set $I_1(x_0) \defeq y_{\mathrm{min}}$, and $I_2(x_0) \defeq y_{\mathrm{max}}$; or $I(x_0) \defeq \emptyset$, in case there is no solution. Finally, we define $I_1(x_k) = I_2(x_k) = y_k$, for all $k \in [n]$, as 
the outputs are noise-free, that is $y_k = f_*(x_k)$, for $k \in [n]$.
{\renewcommand{\arraystretch}{1.3} 
\begin{table}[!t]
\centering
\caption{\vspace*{-4mm}}
\begin{tabular}{|cl|}
\hline
\multicolumn{2}{|c|}{\textsc{Pseudocode: {\newpart Confidence} interval for the noise-free case}} \\ \hline\hline
{\em Input:} & Data sample $\{(x_k, y_k)\}_{k=1}^{n}$, input query point $x_0 \in \CD$,\\
& and risk probability $\alpha \in (0,1)$.\\
{\em Output:} & {\newpart The endpoints of the confidence interval $[\hspace{0.3mm}I_1(x_0), I_2(x_0)\hspace{0.3mm}]$}\\
			  & {\newpart which has confidence probability at least $1-\alpha$.}\\[0.5mm]
\hline \hline
1. & If $x_0 = x_k$ for any $k \in [n]$, return 
$I_1(x_0) = I_2(x_0) = y_k$.\\
2.& Calculate $\kappa \defeq \frac{1}{n} \sum_{k=1}^n y_k^2 + \sqrt{\frac{\log(\alpha)}{-2n}} + \delta_0$. \\ 
3. & Create the extended Gram matrix\\
& $\gramnplus(i+1, j+1)\defeq k(x_i,x_j),$ for $i,j=0,1,...,n$. \\ 
4.& Calculate $\gramnplus^{-1}$ and partition it as:\\ 
& 
$
\begin{bmatrix}
\; c & b\tr\\
\; b & A
\,\end{bmatrix} \defeq\, \gramnplus^{-1}
$\\
5. & Solve the quadratic equation $a_0 y_0^2 + b_0 y_0 + c_0 = 0$, \\ 
& where  $a_0 \defeq c$, $b_0 \defeq 2b\tr y$ and $c_0 = y\tr\hspace{-0.3mm} A y - \kappa$.\\
6. & If there is no solution, return $I(x_0) \defeq \emptyset$; otherwise return\\
& $I_1(x_0) \defeq y_{\mathrm{min}}$, and $I_2(x_0) \defeq y_{\mathrm{max}}$, where $y_{\mathrm{min}} \leq y_{\mathrm{max}}$\\
& are the solutions (which are allowed to coincide).\\[0.5mm]
\hline
\end{tabular}
\label{table:pseudo-noise-free}
\vspace*{-4mm}
\end{table}}

Table \ref{table:pseudo-noise-free} summarizes the proposed algorithm for the case without measurement noise. By observing that if $\kappa$ satisfies $\norm{f}_{\CH}^2 \leq \kappa$, which has probability at least $1-\alpha$, then the construction guarantees that $\mathrm{graph}_{\CD}(f_*) \subseteq \mathcal{I}$, as the region contains all outputs that can be interpolated with a function from $\CH$ which also interpolates the original dataset and 
has norm square at most $\kappa$. Hence, we can conclude that
\medskip

\begin{theorem}{\em Assume that A\ref{A0}, A\ref{A2}, A\ref{A3} and $y_k = f_*(x_k)$, for $k \in [n]$, are satisfied. Let $\alpha \in (0,1)$ be a 
risk probability.
Then, the {\newpart confidence} band of Algorithm \ref{table:pseudo-noise-free} guarantees
$$\mathbb{P}(\, \mathrm{graph}_{\CD}(f_*) \subseteq \mathcal{I}\,) \, \geq \, 1-\alpha.\vspace*{0.8mm}$$}
\end{theorem}
\section{{\newpart Confidence Bands} with Measurement Noise}
Now, we turn to the general case, when the observations of $f_*$ are affected by {\em noises}{\newpart ,} $y_k = f_*(x_k) + \varepsilon_k$, for $k \in [n]$. 

Since now we do not have exact knowledge of the function values at the sample inputs, we cannot directly apply our previous approach. The main idea in this case is that first we need to construct {\em interval estimates} of $f_*$ at some {\em {\newpart observed} inputs}, $\{x_k\}$, which then can be used to bound the norm and to build {\newpart confidence} intervals for the {\em\newpart unobserved} inputs.
\subsection{{\newpart Confidence} Intervals at the {\newpart  Observed} Inputs}
\label{sec:SPS}
We employ the {\em kernel gradient perturbation} (KGP) method, proposed in \cite{csaji2019distribution}, to build {\em non-asymptotically} guaranteed, {\em distribution-free} {\newpart confidence} intervals for $f_*$ at some of the {\em observed} inputs. The KGP algorithm is based on ideas 
from {\em finite-sample system identification} \cite{Algo2018}, particularly, it is an extension of the {\em Sign-Perturbed Sums} (SPS) method \cite{csaji2014sign}.

The KGP method can build non-asymptotically guaranteed distribution-free confidence regions for the RKHS coefficients of the {\em ideal} representation (w.r.t.\ given input points)
of $f_*$.
A representation $f \in \CH$ is called ideal w.r.t.\ $\{x_k\}_{k=1}^{d}$, if it has the property that $f(x_k) = f_*(x_k)$, for all $k \in [\hspace{0.3mm}d\hspace{0.5mm}]$.

{\newpart 
The KGP construction guarantees \cite[Theorem 2]{csaji2019distribution} that the confidence set contains the coefficients of an ideal representation w.r.t.\ $\{x_k\}_{k=1}^{d}$ {\em exactly} with a user-chosen confidence probability, assuming the noises satisfy regularity conditions, e.g., they are symmetric and independent (cf.\ A\ref{A0} and A\ref{A1}).
	
Note that KGP regions are only guaranteed at the {\em observed} inputs. KGP cannot provide confidence bands directly.}

The KGP approach can be used together with a number of kernel methods, such {\newpart as} support vector regression and kernelized LASSO. Here, we use it with {\em kernel ridge regression} (KRR) {\newpart which} is the kernelized version of Tikhonov regularized least squares (LS). It solves the following problem:
\begin{equation}
\label{krr:objective}
\hat{f}_{\scriptscriptstyle\text{KRR}} \; \defeq \; \argmin_{f \in \RKHS}\, \frac{1}{n}\,\sum_{k=1}^n w_i (y_k - f(x_k))^2 \,+\, \lambda\, \| f \|^2_{\RKHS},
\vspace{1mm}
\end{equation}
where $\lambda > 0$, $w_k > 0$, $i \in [n]$, are given (constant) weights.

Using the {\newpart  representer theorem} \cite{hofmann2008kernel} and the reproducing property, the objective of \eqref{krr:objective} can be rewritten as  \cite{csaji2019distribution}\vspace{-0.5mm}
\begin{equation}
\label{krr:obj2}
\frac{1}{n}\,(y - \Ker\hspace{0.2mm} \theta)\tr W (y - \Ker\hspace{0.2mm} \theta) \,+\, \lambda\, \theta\tr \hspace{-0.3mm}\Ker\hspace{0.2mm} \theta,
\end{equation}
where 
$W \doteq \mbox{diag}(w_1,\dots, w_n)$, $K$ is the Gramian matrix,
and $\theta  = (\theta_1, \dots, \theta_n)$ are the 
coefficients of the solution.

Minimizing \eqref{krr:obj2} can be further reformulated as a canonical {\em ordinary least squares} (OLS) problem, $\|\hspace{0.3mm}{\newpart v} \,-\, \Phi\hspace{0.2mm} \theta\hspace{0.3mm}\|^2$, by using\vspace{-0.5mm}
\begin{equation*}
\Phi\, =\, \left[ 
\begin{array}{c}
\,(\nicefrac{1}{\sqrt{n}})\,W^{\frac{1}{2}} \Ker\, \\[1mm]
\sqrt{\lambda}\, \Ker^{\frac{1}{2}}
\end{array}  
\right]\!,\quad
{\newpart v} \,=\, \left[ 
\begin{array}{c}\,
(\nicefrac{1}{\sqrt{n}})\, W^{\frac{1}{2}} y\, \\[1mm]
\;0_n\;
\end{array}  
\right]\!,
\end{equation*}
where $W^{\frac{1}{2}}$ and $\Ker^{\frac{1}{2}}$ denote the principal, non-negative square roots of matrices  $W$ and $\Ker$, respectively. Note that the square roots exist as these matrices are positive semi-definite.

For convex quadratic problems (such as KRR) and {\em symmetric} noises (cf.\ A\ref{A1}), the KGP confidence regions coincide with SPS regions. 
They are {\em star convex} with the LS estimate, $\hat{\theta}$, as a star center. Furthermore, they have {\em ellipsoidal outer approximations}, that is there are regions of the form 
\vspace{-0.5mm}
\begin{equation}
\widehat{\Theta}_{\beta} \; \defeq \; \Big\{\, \theta \in \mathbb{R}^n\, :\, (\theta-\hat{\theta})^\mathrm{T}\frac{1}{n}\Phi\tr\Phi\hspace{0.3mm}(\theta-\hat{\theta})\,\leq\, r \, \Big\},
\end{equation}
where $1-\beta \in (0,1)$ is a given confidence probability \cite{csaji2014sign}. 
The radius of this confidence ellipsoid, $r$, can be computed 
by 
{\em semi-definite programming}:
see \cite[{\newpart Section VI.B}]{csaji2014sign}.

Hence, the construction guarantees $\mathbb{P}(\hspace{0.3mm}\tilde{\theta} \in \Theta_\beta\hspace{0.3mm}) \geq 1-\beta$, where $\tilde{\theta}$ is the coefficient vector of an {\em ideal} representation:
\vspace{-1mm}
$$
\sum_{i=1}^n \tilde{\theta}_i k(x_i, x_k) \,=\, f_*(x_k),
$$
for $k \in [n]$. By defining $\varphi_k \defeq (k(x_1,x_k), \dots, k(x_n,x_k))\tr$, we know that $f_*(x_k) = \varphi_k\tr\tilde{\theta}$, but of course $\tilde{\theta}$ is unknown.

Since $\tilde{\theta}$ is inside the ellipsoid $\widehat{\Theta}_{\beta}$ with probability $\geq 1-\beta$, we could construct (probabilistic) upper and lower bounds of $f_*(x_k)$ by maximizing and minimizing $\varphi_k\tr\theta$, for $\theta \in \widehat{\Theta}_{\beta}$.

These problems (linear objective and ellipsoid constraint) have known solutions: the minimum and the maximum are
$$
\nu_k = \varphi_k\tr\hat{\theta} - (\varphi_k\tr P\varphi_k)^{\frac{1}{2}}, \qquad \mu_k = \varphi_k\tr\hat{\theta} + (\varphi_k\tr P\varphi_k)^{\frac{1}{2}},
$$
where $P = (nr)^{-1} \Phi\tr\Phi$, and $\hat{\theta}$ is the center of the ellipsoid, i.e.,
the {\newpart solution} of the OLS formulation $\|\hspace{0.3mm}{\newpart v} \,-\, \Phi\hspace{0.2mm} \theta\hspace{0.3mm}\|^2$. 

{\newpart  Due to the construction of KGP confidence regions, there is a (extremely small, but nonzero) probability of getting an empty region. In this case, we define $\nu_k = 1$ and $\mu_k = -1$, for all $k \in [n]$. That is, we give an {\em empty interval} for each $f(x_k)$, using a similar representation as in Section \ref{sec:objectives}.}

Finally, we introduced a slight modification to this construction. We can also construct confidence intervals just for the first $d \leq n$ observations by redefining objective \eqref{krr:obj2} as
\begin{equation*}
\frac{1}{n}\,(y - \Ker_1\hspace{0.2mm} \theta)\tr W (y - \Ker_1\hspace{0.2mm} \theta) \,+\, \lambda\, \theta\tr \hspace{-0.3mm}\Ker_2\hspace{0.2mm} \theta,
\end{equation*}
where $K_1 \in \RR^{n \times d}$ is $K$ having the last $n-d$ columns removed, and $K_2\in \RR^{d \times d}$ is $K_1$ having the last $n-d$ rows removed. Hence, we search for $\tilde{\theta} \in \RR^d$ ideal vector, such that 
for $k \in [\hspace{0.3mm}d\hspace{0.5mm}]$, we have $(K_1 \tilde{\theta})(k)=  f_*(x_k)$. 
For the error computation we still use {\em all} measurements ($K_1$ still has $n$ rows). {\newpart It is important that in this case only the first $d$ residuals are perturbed in the construction of the KGP ellipsoid. This}
usually considerably reduces the sizes of the intervals, but then we only have guarantees {\newpart at $d\leq n$ observed inputs}.
\subsection{Bounding the Norm with Measurement Noise}
In the previous section, we built {\em simultaneous} confidence intervals at the sample inputs for the first $d\leq n$ observations, $[\hspace{0.3mm}\nu_k, \mu_k\hspace{0.3mm}]$, for $k \in [\hspace{0.3mm}d\hspace{0.5mm}]$; that is, they have the property
\vspace{-0.2mm}
\begin{equation}
\label{eq:sym.conf.int}
\mathbb{P}\big(\hspace{0.3mm} \forall \hspace{0.3mm}k \in [\hspace{0.3mm}d\hspace{0.5mm}]: f_*(x_k) \in [\hspace{0.3mm}\nu_k, \mu_k\hspace{0.3mm}]\hspace{0.3mm}\big)\, \geq\, 1 - \beta,
\end{equation}
for some (user-chosen) risk probability $\beta \in (0,1)$.

Recall that by Lemma \ref{lemma:Hoeffding.noiseless}, for any $n$, the variable 
\begin{equation}
\label{eq:Hoeffdieng}
\kappa \, \defeq \frac{1}{n} \sum_{k=1}^n f^2_*(x_k) + \sqrt{\frac{\ln(\alpha)}{-2n}} + 
\delta_0,
\end{equation}
is an upper bound of $\norm{f_*}_{\CH}^2$ with probability at least $1-\alpha$.

Using property \eqref{eq:sym.conf.int}, we also know that\vspace{-1mm}
\begin{equation}
\label{eq:sum.max.nu.mu.square}
\sum_{k=1}^{d} f_*^2(x_k)  \,\leq\, \sum_{k=1}^{d} \max\{\nu_k^2, \mu_k^2\},
\end{equation}
with probability at least $1-\beta$. By combining property \eqref{eq:sym.conf.int}, formulas \eqref{eq:Hoeffdieng} and \eqref{eq:sum.max.nu.mu.square},
the results of Lemma \ref{lemma:Hoeffding.noiseless}, as well as using Boole's inequality (the union bound), we have
\medskip
\begin{lemma}
\label{lemma:Hoeffding.noisy}
{\em Assume that A\ref{A0}, A\ref{A2}, A\ref{A3} hold and that confidence intervals 
$[\hspace{0.3mm}\nu_k, \mu_k\hspace{0.3mm}]$, for $k \in [\hspace{0.3mm}d\hspace{0.5mm}]$, satisfy \eqref{eq:sym.conf.int}. 
{\newpart Then,
$$
\mathbb{P}\big(\norm{f_*}_{\CH}^2 \leq \tau \hspace{0.3mm}\big)\,  \geq \,1-\alpha-\beta,
$$
with the following choice of the upper bound $\tau$:
$$
\tau \, \defeq\, \frac{1}{d} \sum_{k=1}^{d} \max\{\nu^2_k,\mu^2_k \} + \sqrt{\frac{\ln(\alpha)}{-2d}} + 
\delta_0.$$
}}
\vspace{0mm}
\end{lemma}
\begin{remark}
Although we only used the first $d$ observations for estimating the norm (square), the intervals $[\hspace{0.3mm}\nu_k, \mu_k\hspace{0.3mm}]$, for $k \in [\hspace{0.3mm}d\hspace{0.5mm}]$, incorporate information about the {\em whole} sample. 
{\newpart The ``optimal'' choice of $d$ leading to small intervals is an open question, 
in practice $d = \mathcal{O}(\sqrt{n})$ often works well.}
\end{remark}

\subsection{Interval Endpoints with Measurement Noise}
The final step is to construct a {\newpart confidence} interval for a given input {\em query point} $x_0 \in \CD$ with $x_0 \neq x_k$, for $k \in [\hspace{0.3mm}d\hspace{0.5mm}]$.

We extend the Gram matrix with query point $x_0$, 
$$
\gramdplus({i+1},{j+1})\, \defeq \, k(x_i,x_j),
$$
for $i, j = 0,1, \dots ,d$; but we only use the first $d$ data points.

{\renewcommand{\arraystretch}{1.3} 
\begin{table}[!t]
\centering
\caption{\vspace*{-4mm}}
\begin{tabular}{|cl|}
\hline
\multicolumn{2}{|c|}{\textsc{Pseudocode: {\newpart Confidence} interval with measurement noise}} \\ \hline\hline
{\em Input:} & Data sample $\{(x_k, y_k)\}_{k=1}^{n}$, input query point $x_0 \in \CD$,\\
& risk probabilities $\alpha \in (0,1)$ and $\beta \in (0,1)$.\\
{\em Output:} & {\newpart The endpoints of the confidence interval $[\hspace{0.3mm}I_1(x_0), I_2(x_0)\hspace{0.3mm}]$}\\
& {\newpart which has confidence probability at least $1-\alpha-\beta$.}\\[0.5mm]
\hline \hline
1. & Select $d \in [n]$, the number of confidence intervals built for\\
& a subset of {\newpart observed} inputs. Default choice: $d = \ceil{\sqrt{n}\hspace{0.3mm}}$. \\
2.& Construct $1-\beta$ level simultaneous confidence intervals for\\
& $\{f_*(x_k)\}_{k=1}^{d}$, that is $[\hspace{0.3mm}\nu_k, \mu_k\hspace{0.3mm}]$, for $k \in [\hspace{0.3mm}d\hspace{0.5mm}]$, with \eqref{eq:sym.conf.int}.\\
 & (e.g., apply the KGP method discussed in Section \ref{sec:SPS})\\
3.& Set $\tau \, \defeq\, \frac{1}{d} \sum_{k=1}^{d} \max\{\nu_k^2, \mu_k^2 \} + \sqrt{\frac{\ln(\alpha)}{-2d}} + 
\delta_0$. \\ 
4. & Solve both convex optimization problems given by \eqref{noisy-opt-min-max}.\\
5. & If there is no solution, return $I(x_0) \defeq \emptyset$; otherwise return\\
& $I_1(x_0) \defeq z_{\mathrm{min}}$ and $I_2(x_0) \defeq z_{\mathrm{max}}$, where $z_{\mathrm{min}} \leq z_{\mathrm{max}}$\\
& are the solutions (which are allowed to coincide).\\[0.5mm]
\hline
\end{tabular}
\label{table:pseudo-noisy}
\vspace*{-4mm}
\end{table}}

We have to be careful with the optimization problems, as now we do not know the exact function values, we only have potential intervals for them. Therefore, all function values are treated as decision-variables, which can take values from the given confidence intervals. Hence, we have to solve
\begin{equation}
\label{noisy-opt-min-max}
\begin{split}
\mbox{min\,/\,max} &\quad z_{0} \\[0.5mm]
\mbox{subject to} &\quad (z_0, \dots, z_d) \gramdplus^{-1} (z_0, \dots, z_d)\tr \leq\, \tau\\
&\quad \nu_1 \leq z_1 \leq \mu_1,\; \dots,\; \nu_d \leq z_d \leq \mu_d\\[0.5mm]
\end{split}
\end{equation}
where ``min\,/\,max'' again means that the problem have to be solved as a minimization and as a maximization (separately). 

These problems are {\em convex}, therefore, they can be solved efficiently.
The optimal values, denoted by $z_{\mathrm{min}}$ and $z_{\mathrm{max}}$, are the {\em endpoints} of the {\newpart confidence} interval: 
$I_1(x_0) \defeq z_{\mathrm{min}}$, and $I_2(x_0) \defeq z_{\mathrm{max}}$. 
If \eqref{noisy-opt-min-max} is infeasible, e.g., we get an empty KGP ellipsoid, we set $I(x_0) = \emptyset$, i.e., we use $I(x_0) =(\hspace{0.3mm}1, -1\hspace{0.3mm})$.

Table \ref{table:pseudo-noisy} summarizes the algorithm to construct the endpoints of a confidence interval at a given query point, in case of having measurement noises. Its theoretical guarantee is:

\medskip
\begin{theorem}{\em Assume that A\ref{A0}, A\ref{A1}, A\ref{A2}, A\ref{A3} are satisfied. Let $\alpha, \beta \in (0,1)$ be given risk probabilities.
Then, the confidence band built by Algorithm \ref{table:pseudo-noisy} described above guarantees
$$\mathbb{P}(\, \mathrm{graph}_{\CD}(f_*) \subseteq \mathcal{I}\,) \, \geq \, 1-\alpha - \beta.$$}
\end{theorem}
\vspace{4mm}
\begin{remark}
Applying the KGP approach in Algorithm \ref{table:pseudo-noisy} is optional. One could use any other construction that provides simultaneous confidence intervals for a subset of $\{f_*(x_k)\}$, cf.\ \eqref{eq:sym.conf.int}. Another approach could be to assume sub-Gaussian or sub-exponential noises and use their tail bounds to ensure \eqref{eq:sym.conf.int}.
\end{remark}

\begin{figure}[!t]
    \centering
	\hspace*{-2mm}	
	\includegraphics[width = 1.02\columnwidth]{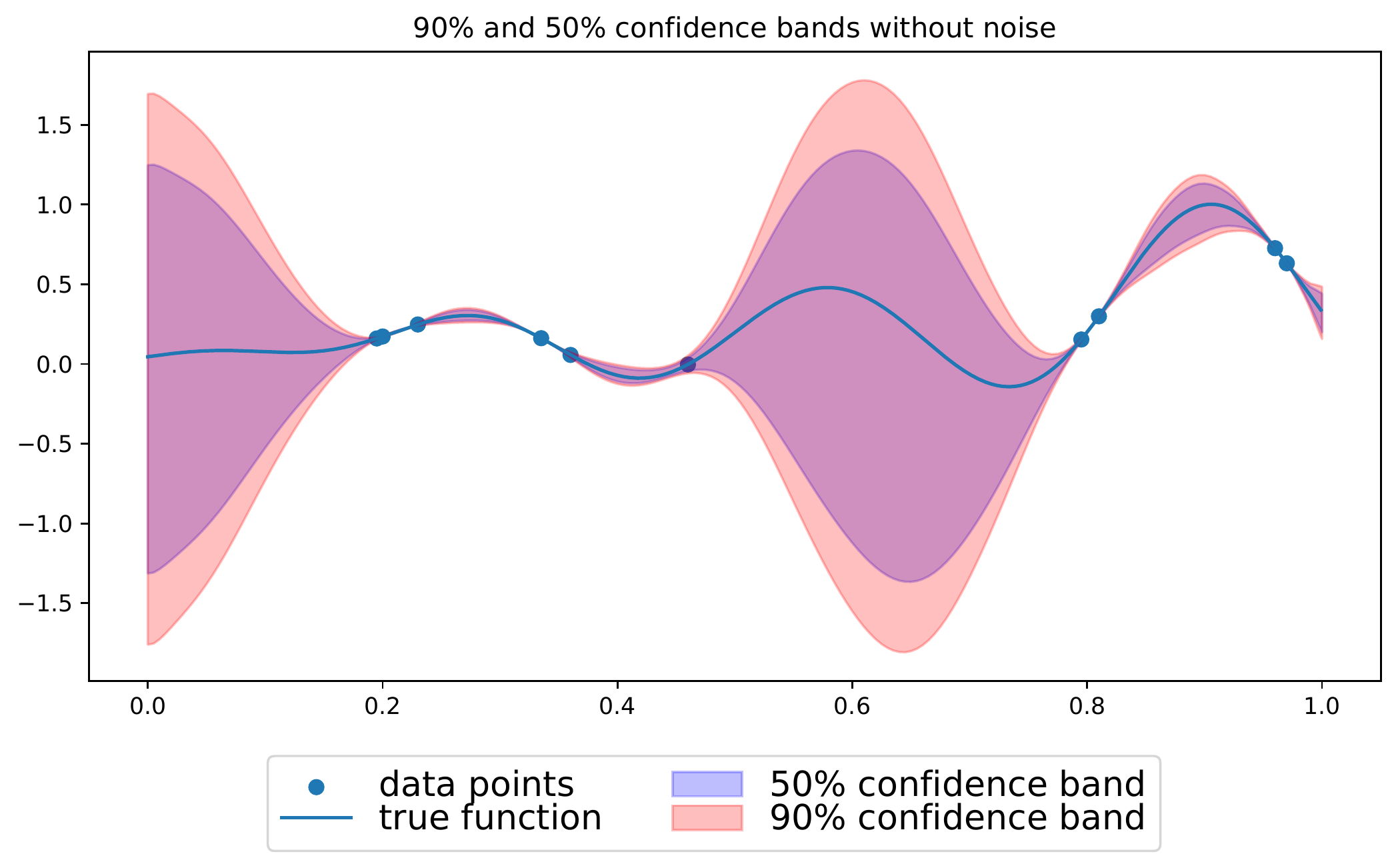} 	
    \caption{Nonparametric {\newpart  confidence bands} for the noise-free setting.}
\label{fig:experiment1}
\end{figure} 

\begin{figure}[!t]
    \centering
	\hspace*{-2mm}	
	\includegraphics[width = 1.02\columnwidth]{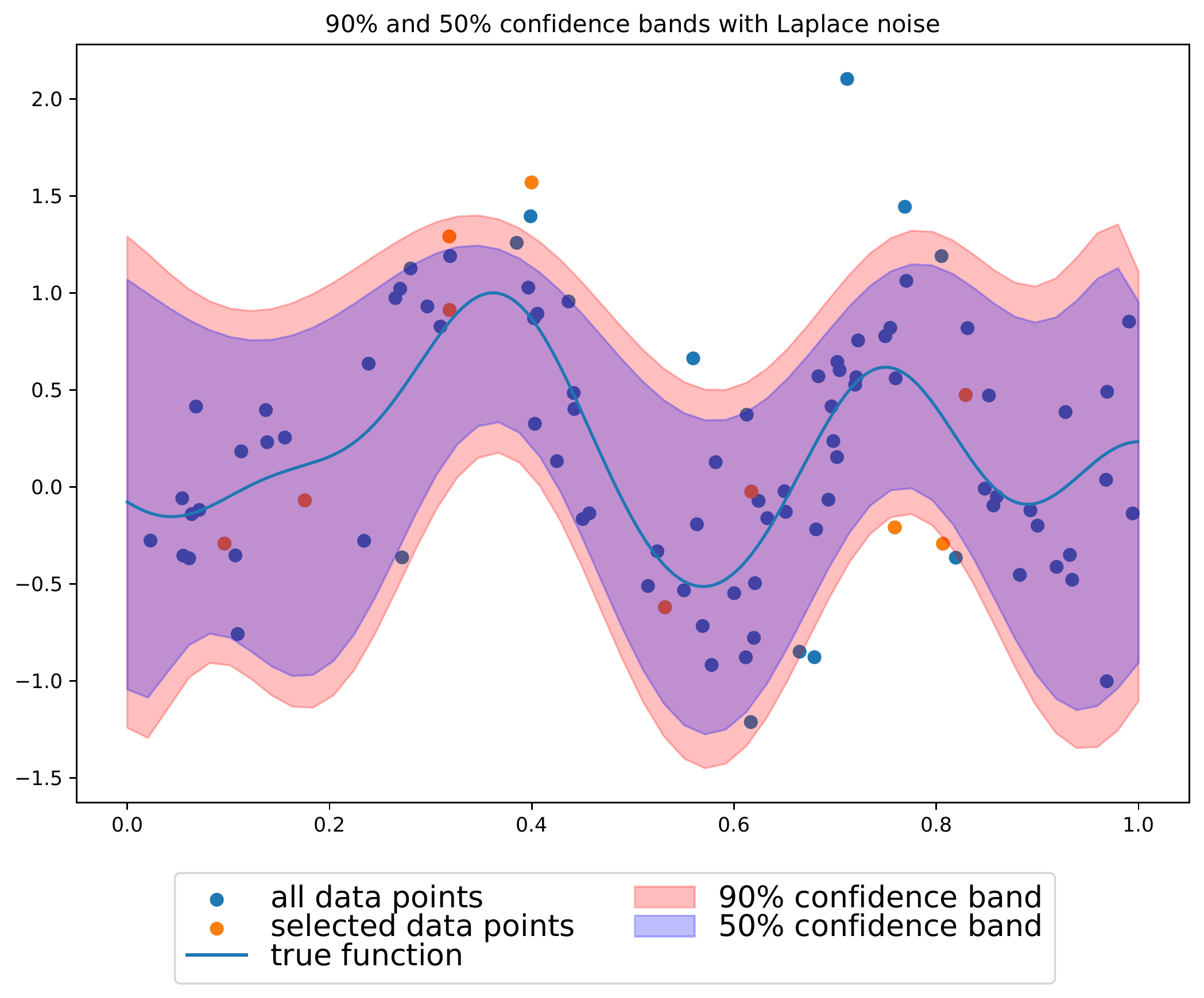}
    \caption{Nonparametric {\newpart  confidence bands} with measurement noise.}
\label{fig:experiment2}
\vspace*{-2mm}
\end{figure}
\section{Numerical Experiments}
The algorithms were also tested numerically.
We used a Paley-Wiener RKHS with $\eta = 30$. The ``true'' 
function was constructed as follows: first, $20$ random input points $\{\bar{x}_k\}_{k=1}^{20}$ were generated, with uniform distribution on $[\hspace{0.3mm}0,1]$. Then $f_*(x) = \sum_{k=1}^{20} w_k k(x, \bar{x}_k)$ was created, where each $w_k$ had a uniform distribution on $[-1,1]$. The function was normalized, in case its maximum exceeded $1$. Then, $n$ random observations were generated about $f_*$. In the noisy case, 
$\{\varepsilon_k\}$ had Laplace distribution with location $\mu = 0$ and scale $b = 0.4$ parameters.

In the noise-free case, we used $n=10$ observations, and created confidence bands with risk $\alpha = 0.1$ and $0.5$. Figure \ref{fig:experiment1} demonstrates that 
in the noise-free setting a very small sample size can lead to informative nonparametric confidence bands.

In case of measurement noises, $n=100$ sample size was used with $d=20$ (orange points). {\newpart Confidence bands} with risk $\alpha + \beta = 0.1$ and $0.5$ are illustrated in Figure \ref{fig:experiment2}. We simply used $\alpha = \beta$ in these cases. The results indicated that even with limited information,
adequate regions can be created.

\section{Conclusions}

In this paper a nonparametric and distribution-free {\newpart  method was introduced to build simultaneous confidence bands for bounded, band-limited functions}. The construction was first presented for the case when there are no measurement noises, then it was extended allowing symmetric noises. Besides having non-asymptotic theoretical guarantees, the approach was also demonstrated numerically, supporting its feasibility. 

\bibliographystyle{ieeetr}
\bibliography{references} 

\end{document}